\documentclass[runningheads]{llncs}
\usepackage{graphicx}
\usepackage{xcolor}
\usepackage{subcaption}
\usepackage[ruled,vlined]{algorithm2e}
\include{pythonlisting}
\SetArgSty{textnormal}
\usepackage{fancyhdr,graphicx,amsmath,amssymb}
\usepackage{hyperref}
\hypersetup{
  citecolor=blue,
  colorlinks=true,
  linkcolor=blue,
}
\begin{document}
\title{Recovering the Imperfect: Cell Segmentation in the Presence of Dynamically Localized Proteins}
\titlerunning{Cell Segmentation in the Presence of Dynamically Localized Proteins}
\author{\"Ozg\"un \c{C}i\c{c}ek\textsuperscript{*} \inst{1}  \and
Yassine Marrakchi\textsuperscript{*} \inst{1,2} \and 
Enoch Boasiako Antwi \inst{1,2,3} \and \\
Barbara Di Ventura \inst{1,2} \and 
Thomas Brox \inst{1,2}}
\institute{
University of Freiburg, Germany
\and Signalling Research Centres BIOSS and CIBSS, Freiburg, Germany
\and Heidelberg Biosciences International Graduate School (HBIGS), Germany
\email{\{cicek,marrakch\}@cs.uni-freiburg.de}
}
\authorrunning{\c{C}i\c{c}ek et al.}
\maketitle              
\begin{abstract}
Deploying off-the-shelf segmentation networks on biomedical data has become common practice, yet if structures of interest in an image sequence are visible only temporarily, existing frame-by-frame methods fail. In this paper, we provide a solution to segmentation of imperfect data through time based on temporal propagation and uncertainty estimation. We integrate uncertainty estimation into Mask R-CNN network and propagate motion-corrected segmentation masks from frames with low uncertainty to those frames with high uncertainty to handle temporary loss of signal for segmentation. We demonstrate the value of this approach over frame-by-frame segmentation and regular temporal propagation on data from human embryonic kidney (HEK293T) cells transiently transfected with a fluorescent protein that moves in and out of the nucleus over time. The method presented here will empower microscopic experiments aimed at understanding molecular and cellular function.
\end{abstract}
{
\renewcommand{\thefootnote}{\fnsymbol{footnote}}
\footnotetext[1]{equal contribution}
}
\section{Introduction}
\begin{figure}[t]
\centering
\begin{subfigure}{.3\textwidth}
  \centering
  \includegraphics[width=0.35\linewidth]{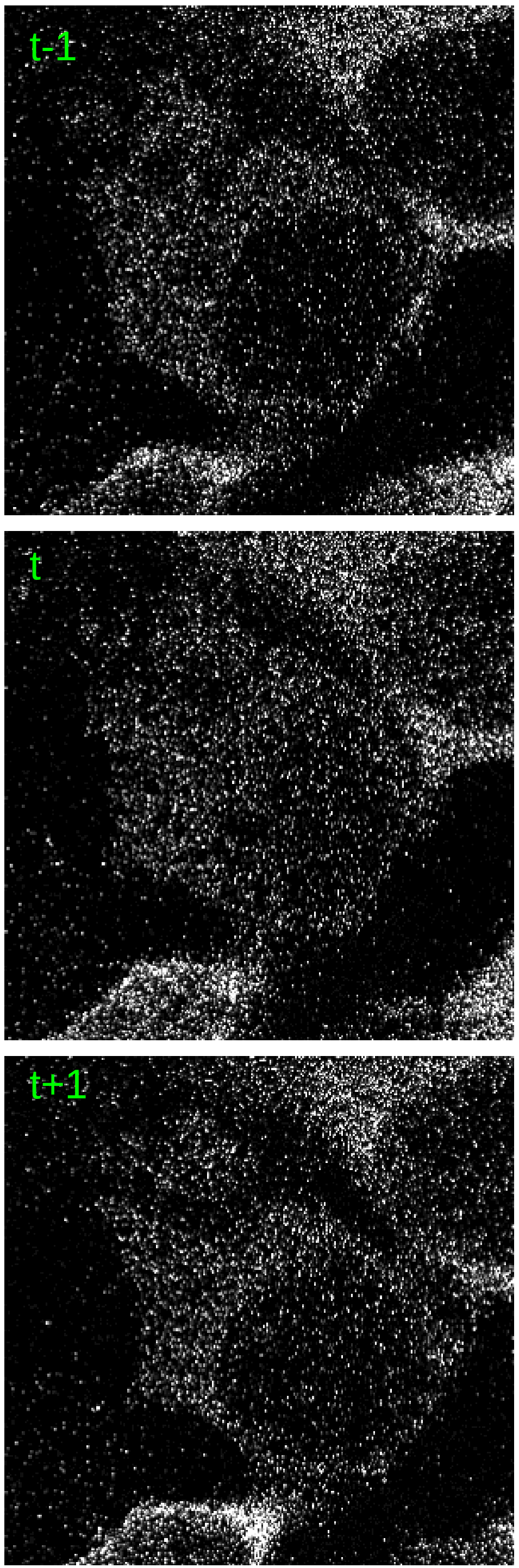}
  \caption{}
  \label{fig:sub1}
\end{subfigure}%
\begin{subfigure}{.7\textwidth}
  \centering
  \includegraphics[width=0.75\linewidth]{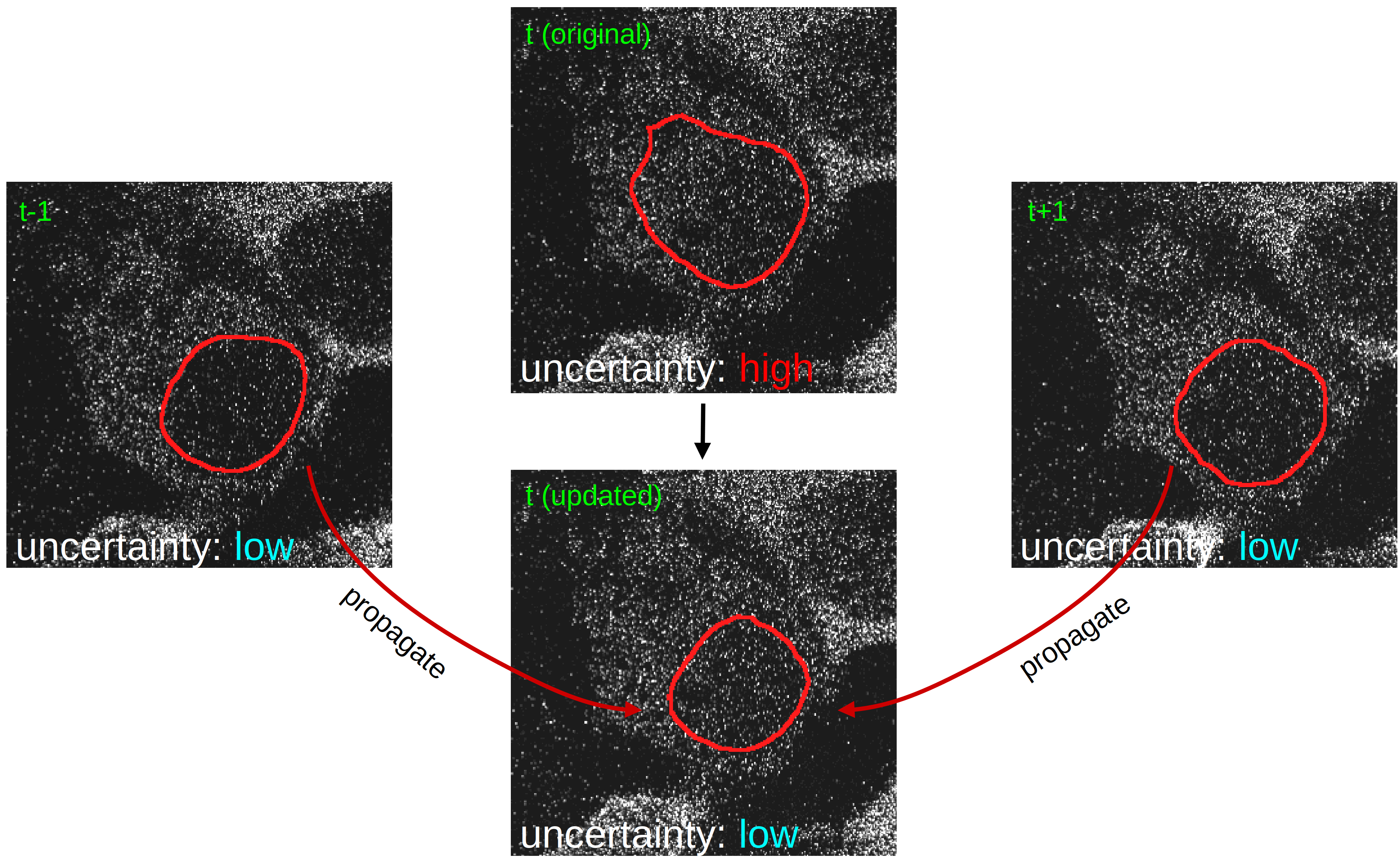}
  \caption{}
  \label{fig:sub2}
\end{subfigure}
\caption{(a) Exemplary time-lapse images of HEK293T cells depicting an oscillatory nuclear signal. (b) Oscillation at time $t$ causing bad nuclei segmentation (up) and the corrected segmentation of it using our propagation method (down).}
   \label{fig:teaser}
\end{figure}

In the past decades, it has become evident that proteins are very dynamic and their localization within the cell dictates which function they perform~\cite{1}. Cell biologists carry out time-lapse fluorescence microscopy experiments to study protein localization and unravel how this affects the protein's function. Optogenetics can be used to control the localization of the protein of interest~\cite{5} to observe the effects of dynamic localization patterns. 
Take Figure~\ref{fig:teaser} for an example: HEK293T cells expressing mCherry fused to the optogenetic tool LINuS~\cite{linus} were observed over time to analyze the effect of blue light on the nuclear import of the protein. In the absence of blue light, the fusion protein localizes predominantly in the cytosol. Hence, the nucleus appears dark. However, in presence of blue light, the fusion protein enters the nucleus making it appear bright and difficult to distinguish from the cytosol. Giving light repeatedly creates oscillations of the protein in and out of the nucleus in time. 

Learning-based methods, such as U-Net~\cite{unet} and Mask R-CNN~\cite{maskrcnn}, succeed at segmenting structures in data with clearly visible patterns, but fail when the visibility deteriorates. When the signal in the nucleus is similar to that in the cytosol, nuclei segmentation from any network is not reliable. While in a single image it is not possible to improve these segmentations, past and future frames in a video provide additional information for refinement. An expert can play the video back and forth to infer the segmentation of ambiguous nuclei. An automated segmentation method, too, must (1) automatically identify critical frames and (2) propagate predictions from neighbouring frames. 

In this paper, we address both challenges by (1) equipping Mask R-CNN with uncertainty estimation to identify erroneous predictions and (2) incorporating optical flow to improve the identified erroneous predictions by propagating certain predictions from neighboring frames. Doing so, we introduce the most recent uncertainty estimation methods in biomedical instance segmentation and solve a real task commonly experienced in signalling studies which is not yet addressed. So far, nuclear markers have been employed in the experiments so that the available automated segmentation tools can be used~\cite{3}. However, additional markers cause unreliable quantification since different proteins bleed-through and interfere with each other. They also limit the channel space needed for other proteins of interest. The presented method makes the use of a nuclear marker dispensable.
 
\section{Related Work}
We are the first to address instance segmentation of structures with an oscillatory fluorescent signal in biomedical videos. Our work is related to methods that benefit from temporal features in their design. Milan et al.~\cite{rnntracking} and Payer et al.~\cite{ctc1} used Recurrent Neural Network (RNN) to aggregate temporal features. Paul et al.~\cite{segpropflow} incorporated temporal cues for segmentation via optical flow. Similarly, Jain et al.~\cite{segpropflow2} propagated features instead of segmentation with flow from key frames. A similar idea was applied to instance segmentation by Bertasius and Torresani~\cite{maskrcnnprop}. 
Although these methods seem close, the task is different: (1) they heavily rely on dense annotation in time to learn interpolations explicitly while we cannot afford it due to high cost of expert annotation and difficulty in fine-grained annotation of imperfect frames and (2) they benchmark only on visible objects, while we are solely interested in objects with limited visibility.

 \begin{figure}[t]
   \centering
     \includegraphics[width=1\textwidth]{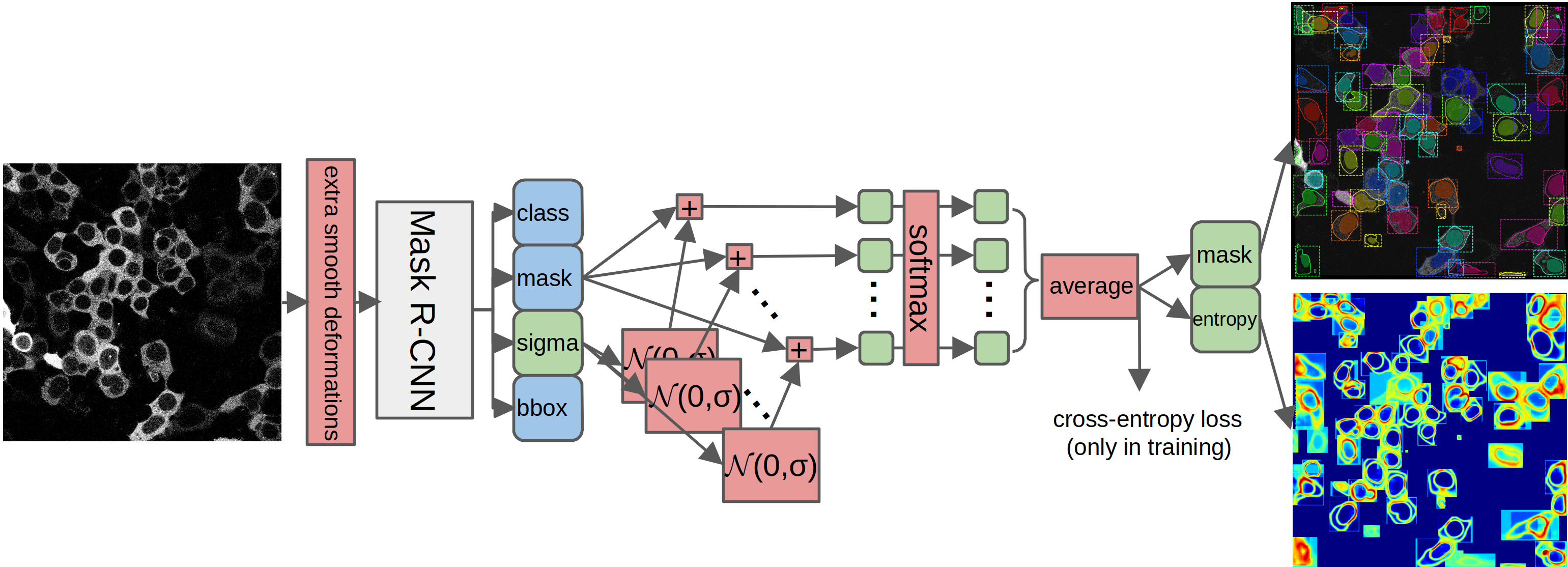}
   \caption{Overview of Mask R-CNN with added data uncertainty. Changes to the original architecture are shown in red (operations) and green (outputs).}
   \label{fig:overview}
 \end{figure}

\section{Methods}
\subsection{Instance-Aware Segmentation with Uncertainty Estimation}
We base our model on Mask R-CNN~\cite{maskrcnn} equipped with elastic deformations of U-Net~\cite{unet} to create additional biomedically plausible images on-the-fly for better generalization. 
We incorporate uncertainty estimation in our Mask R-CNN architecture to  detect erroneous predictions; see Figure~\ref{fig:overview}.
We consider data uncertainty (aleatoric), model uncertainty (epistemic), and their combination. 

\textbf{Data uncertainty.} For data uncertainty, we use the modified cross-entropy loss~\cite{whichuncertainties} in the mask branch of Mask R-CNN. This models the data uncertainty as the learned noise scale from the data. 
To learn both the class scores and their noise scale, the negative expected log-likelihood is minimized for pixel $i$
as:
\begin{equation} \small{
     L(s)=-\sum_{i}{\log\left[{\frac{1}{T}}\sum_{t}{\exp(\hat{s}_{i,t,c^{'}}-\log{\sum_{c}{\exp(\hat{s}_{i,t,c})}})}\right]}}
     \mathrm{\,,}
     \label{eq:data_uncertainty}
\end{equation}
\noindent where $c^{'}$ is the correct class among all classes ($c$) and $\hat{s}$ are predicted logits corrupted by Gaussian noise with standard deviation $\sigma$. $\sigma$ is also learned by the network alongside with the logits. 
Figure~\ref{fig:overview} shows where this loss is embedded in the network.
At test time, the uncertainty is computed as the entropy of the class pseudo-probabilities for each pixel as:
$\mathcal{U}(p) = -\sum_{c}{p(c)\log{p(c)}}$.

\textbf{Model uncertainty.} Model uncertainty requires sampling from the model. We 
experimented with the latest sampling strategies. \textbf{Dropout} is one of the common techniques to sample from networks~\cite{whichuncertainties}. \textbf{Ensemble} replaces the sampling by dropout by individually trained networks~\cite{ensemble}. \textbf{SGDR ensemble} replaces the individual trained networks by the pre-converged snapshots of the same model~\cite{sgdrensemble}. Pre-converged models are obtained at the end of each SGDR (SGD with warm restarts)~\cite{sgdr} cycle. 
\textbf{WTA} (Winner-Takes-All)~\cite{mhp} aims for training a single network with multiple heads where at each iteration only the head with the best prediction gets penalized. 
\textbf{EWTA} (Evolving WTA)~\cite{ewta} is a variant of WTA which improves the trade-off between diversity and consistency. 

\subsection{Uncertainty-Based Nuclei Propagation}
We base the nuclei propagation on the cell tracker\footnote{\url{http://celltrackingchallenge.net/}} by Ronneberger et al.~\cite{unet}. This overlapping-based tracker is well suited for the data at hand, which does not show large motion over time. Despite its simplicity, it is one of the top performing cell trackers in the challenge. After computing the overlapping-based tracks, we go over all the frames of each track with ascending average nuclei uncertainty order (without nuclei, uncertainty is set to infinity) and update the identified uncertain predictions by propagating masks from more certain neighbours as depicted in Figure~\ref{fig:teaser}. A prediction is marked uncertain if its average uncertainty is higher than a threshold $\theta$. Neighbours are more certain if they meet a relative threshold of $\alpha$ in case of a single-side propagation and $\beta$ in case of a two-sided propagation as detailed in Algorithm~\ref{algo:tracking}. This resembles the procedure that experts follow to find non-visible nuclei by sliding over time. 

\begin{algorithm}[t]
    \caption{{\bf Uncertainty-Based Nuclei Propagation } \label{algo:tracking}}
    \KwIn{instance segmentation masks $\mathcal{S}$, uncertainty maps $\mathcal{U}$, motion per consecutive frame pairs $\mathcal{M}$, hard threshold $\theta$, relative thresholds: $\alpha$,$\beta$}
\KwOut{tracks}
\BlankLine

\nl \text{tracks}$\leftarrow$ IoU-BasedCellTracker($\mathcal{S}$)\;
\nl \For {track \textit{t} in tracks}{
    \nl \For {frame \textit{f} in track \textit{t}}{
            \nl $\bar{u}_{t,f}$ $\leftarrow$ mean\_over\_nuclei($\mathcal{U}_{t,f},S$)\;
        }
        \nl order$\leftarrow$argsort($\bar{u}_{t}$)\;
        \nl \For {frame id \textit{f} in order}{
                \nl \If {$\bar{u}_{t,f}\geq\theta$ and $\beta\times\bar{u}_{t,f} \geq \bar{u}_{t,f-1}$ and $\beta\times\bar{u}_{t,f} \geq \bar{u}_{t,f+1}$}{  
                           \nl $\text{mask}_\text{prev}$ $\leftarrow$ warp($\mathcal{S}_{t,f}$,$\mathcal{S}_{t,f-1}$,$\mathcal{M}$)\; 
                           \nl $\text{mask}_\text{next}$ $\leftarrow$ warp($\mathcal{S}_{t,f}$,$\mathcal{S}_{t,f+1}$,$\mathcal{M}$)\; 
                           \nl $\mathcal{S}_{t,f}$ $\leftarrow$ union($\text{mask}_\text{prev},\text{mask}_\text{next}$)\;
                    }
                \nl \ElseIf {$\bar{u}_{t,f}\geq\theta$ and $\alpha\times\bar{u}_{t,f} \geq \bar{u}_{t,f-1}$}{
                        \nl $\mathcal{S}_{t,f}$ $\leftarrow$ warp($\mathcal{S}_{t,f}$,$\mathcal{S}_{t,f-1}$,$\mathcal{M}$)\;         
                    } 
                \nl \ElseIf {$\bar{u}_{t,f}\geq\theta$ and $\alpha\times\bar{u}_{t,f} \geq \bar{u}_{t,f+1}$}{
                        \nl $\mathcal{S}_{t,f}$ $\leftarrow$ warp($\mathcal{S}_{t,f}$,$\mathcal{S}_{t,f+1}$,$\mathcal{M}$)\;          
                    }             
        }
    }

\end{algorithm}

\subsection{Motion Estimation for Biomedical Videos}
Propagation of predictions over time requires motion estimation between frames to warp the certain predictions onto the less certain ones. One simple way is to warp by the shift and scaling parameters computed between the not yet updated and neighbouring nuclei predictions. This approach assumes the shape of the nuclei does not change over time; however, slight deformations can occur. Optical flow can provide fine-grained motion. Recent optical flow methods for natural images are networks~\cite{fn} trained on synthetic datasets. These methods perform well on real images, but their performance deteriorates as the gap between real and synthetic images grows. Our data is very different from existing synthetic datasets and no synthetic data exists for biomedical data.
Therefore, with the prior knowledge that expected flow in our videos can be well explained by smooth deformations, and being able to generate them on the fly via deformation augmentations of U-Net, we explicitly train a network to predict these deformations as explored by Sokooti et al.~\cite{def_flow}. We warp an image ($I_{t+1}$) backward in time with randomly generated smooth deformations ($f_{t\rightarrow{t+1}}$) to obtain the previous image ($I_{t}$). 
Then we train a FlowNet on the on-the-fly-generated image pairs and ground-truth flows. Even though the fluorescent signal in the nuclei regularly disappears making the optical flow challenging to estimate, the motion of the cytosol helps infer the motion of the nuclei.

\section{Experiments}
\textbf{Implementation details.} We based our implementation on the Mask R-CNN by Abdulla~\cite{matterport}. We incorporated publicly available elastic deformations\footnote{\url{https://github.com/fcalvet/image\_tools}} for U-Net~\cite{unet} as additional augmentation. We used ResNet50-FPN as backbone. We used softmax cross entropy for the mask head with $3$ classes (background, cytosol and nuclei). We trained the networks from scratch for $12k$ epochs except the SGDR ensemble, which was trained for $15k$ with $3k$ cycles. We used the last pre-converged models. The WTA and EWTA models were trained for $11k$ and then merged with a network trained for $4k$. We used $4$ as ensemble size for all methods. Ensembling was performed over bounding boxes. For nuclei propagation, we used $\theta=0.5$, $\alpha=0.7$ and $\beta=0.85$. For Mask R-CNN and FlowNet (variant C) training we used deformations with $3$ and $10$ control points, respectively and deformation magnitude of $10$. Please see the original papers for more details. Our code and examplar dataset is publicly available\footnote{\url{https://lmb.informatik.uni-freiburg.de/Publications/2020/CMB20/}}.
\newline\textbf{Data and annotation.} While the challenge we address is common in signalling studies, there is no public dataset for this purpose. Our data was generated after 24 hours of transient transfection of human embryonic kidney (HEK293T) cells with a construct expressing the fusion protein mCherry-LINuS with a ZEISS LSM780 confocal microscope. MaskRCNN was trained on $82$ cropped images from original $1800x1800$ images with pixel size $0.11x0.11$ and $6$ cells on average in each and tested on $2$ full-sized images with $73$ cells in total to validate instance segmentation and uncertainty estimation. We evaluated our propagation algorithm on an unseen video with $35$ full-sized frames with randomly selected $117$ cells annotated by experts. In cases where the nuclei appeared ambiguous experts interpolated the annotations from past and future frames. 

\subsection{Instance Segmentation and Uncertainty Estimation Evaluation}
We evaluated our instance segmentation by mean average precision (mAP). In Table~\ref{tab:uncertainty_table} we refer to standard mAP as mAP (sm) since it is based on softmax scores. To further evaluate the quality of the uncertainty estimation, we replaced the softmax scores by average entropy over the cell and re-computed mAP (mAP (ent)). This simulates our approach as we rely on averaged uncertainties in our nuclei propagation. This measure shows how reliable the predicted uncertainties are at ranking the prediction quality in the precision-recall curves, which are used commonly to evaluate uncertainty estimation for classification tasks. 
In Table~\ref{tab:uncertainty_table} we see that WTA merged is the best at $0.5$ IoU (Intersection over Union) threshold and at $0.75$ threshold ensemble is the best. In the rest of our experiments we used the WTA merged with data uncertainty since it is computationally more efficient than ensemble.

\begin{table*}[ht]
    \caption{Quantitative evaluation for instance segmentation and uncertainty estimation in mAP (@0.5/@0.75 IoU).
        \label{tab:uncertainty_table}
    }
    \begin{center}
    \resizebox{0.7\textwidth}{!}{%
\begin{tabular}{|l||c|c||c|c|}
    \hline
        & 
        \multicolumn{2}{c||}{model uncertainty} &
        \multicolumn{2}{c|}{combined uncertainty}
        \\
        \cline{2-5}
        &
        mAP (sm) & mAP (ent) &
        mAP (sm) & mAP (ent)
        \\
    \hline
        Single &
        $0.77/0.48$ & $0.80/0.49$ &
        $0.74/0.60$ & $0.83/0.69$
        \\
    \hline
        Dropout &
        $0.74/0.61$ & $0.78/0.65$ &
        $0.77/0.61$ & $0.83/0.67$
        \\
    \hline 
        Ensemble & 
        $0.82/\mathbf{0.64}$ & $0.78/0.61$ &
        $0.78/0.63$ & $0.83/\mathbf{0.70}$
        \\
    \hline 
        SGDR Ensemble & 
        $0.75/0.54$ & $0.72/0.51$ &
        $0.71/0.49$ & $0.63/0.44$
        \\
    \hline 
        WTA Merged & 
        $0.74/0.47$ & $0.82/0.49$ &
        $\mathbf{0.83}/0.56$ & $\mathbf{0.85}/0.64$
        \\        
    \hline
        EWTA Merged & 
        $0.64/0.51$ & $0.73/0.58$ &
        $0.80/\mathbf{0.64}$ & $0.77/0.59$
        \\        
    \hline
\end{tabular}
}

    \end{center} 
\end{table*}

\subsection{Nuclei Propagation Evaluation}
We report the mean IoU of the nuclei segmentation for all experiments in Table~\ref{tab:tracking}. We use \textit{interpolated} for nuclei completely missed before our improvement, \textit{updated} for the nuclei that were segmented but our method decided to improve, \textit{non-updated} for the nuclei that were chosen not to be improved by our method and \textit{all} for all the mentioned cases. The first row shows the results before any propagation algorithm. We present $3$ variants of motion used in propagation (column: \textit{warped with}). We also explored using the certainties as pixel-wise weights in fusing segmentations from candidate neighbors and computed the weighted average to find the final mask. To isolate the gain by our uncertainty-based error detection, we created a baseline which is identical to our mean-flow variant, but performs the propagation on all the nuclei over a track. The significant improvement obtained by all our variants, especially flow variants without fusion, shows that our method can effectively improve erroneous nuclei predictions. The baseline that propagates to all frames (\emph{all}) had a lower performance on the \textit{non-updated} frames, showing that propagation independent of an uncertainty measure harms nuclei with high confidence (approx. $90\%$ of all nuclei).

\begin{table*}[ht]
    \caption{Quantitative evaluation of uncertainty-based propagation in mean IoU for all/updated/interpolated/non-updated nuclei with respective nuclei counts.
        \label{tab:tracking}
    }
    \begin{center}
    \resizebox{0.9\textwidth}{!}{%

\begin{tabular}{|c|c|c||c|c|c|c|}

\hline
\textbf{update} & \textbf{warp with} & \textbf{mask fusion}  & \textbf{all} (117)  & \textbf{updated} (51)  & \textbf{interpolated} (11) & \textbf{non-updated} (55)    \\ \hline

none      & none            & no  & 0.62      & 0.55      & 0.00      & \textbf{0.80}     \\ \hline
uncertain & shift+scale     & no  & 0.71      & 0.68      & 0.39      & \textbf{0.80}     \\ \hline
uncertain & mean flow       & no  & \textbf{0.73}      & 0.71      & \textbf{0.45}      & \textbf{0.80}     \\ \hline
all       & mean flow       & no  & 0.69      & 0.70      & 0.40      & 0.74     \\ \hline
uncertain & pixel-wise flow & no  & \textbf{0.73}      & \textbf{0.72}      & 0.44      & \textbf{0.80}     \\ \hline 
uncertain & pixel-wise flow & yes & 0.72      & 0.70      & 0.40      & \textbf{0.80}     \\ \hline

\end{tabular}
}

    \end{center} 
\end{table*}
\vspace{-1cm}
\section{Qualitative Results}

\subsection{Slight Signal Loss}
\begin{figure}[h!]
   \centering
   \includegraphics[width=0.9\textwidth]{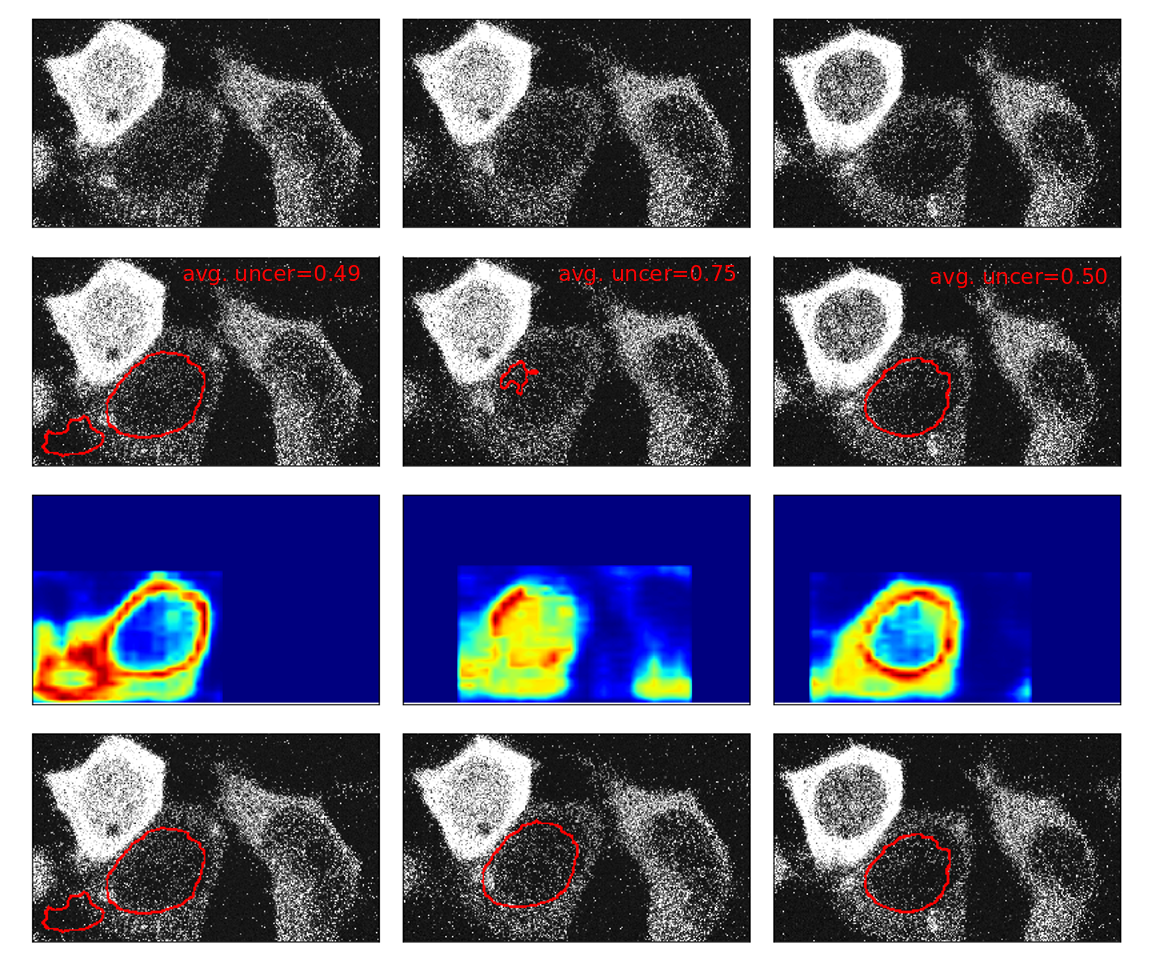}
   \caption{Example of a nucleus getting slightly less visible due to increased noise at time $t$. Columns correspond to time points $t-1$, $t$ and $t+1$, from left to right respectively. In the first row we show the raw images. In the second row we overlay MaskRCNN outputs with the input image and report the average uncertainty. Pixel-wise uncertainty maps are presented in the third row while updated masks are overlaid to the original images in the last row.}
   \label{fig:overview1}
 \end{figure}

Since the network has been trained on a very clean training data, even small shifts in the signal distribution might disturb its performance and cause failures as in Figure~\ref{fig:overview1}. Notice that in the first frame, we have two connected components wrongly building together the nucleus mask. As each cell can only have a single nucleus, we consider propagating one which is the most certain. The remaining connected component can be filtered with a simple post-processing step. We see that our method improves upon the initial segmentation masks significantly.

\subsection{Extreme Signal Loss}
\begin{figure}[h!]
   \centering
     \includegraphics[width=0.9\textwidth]{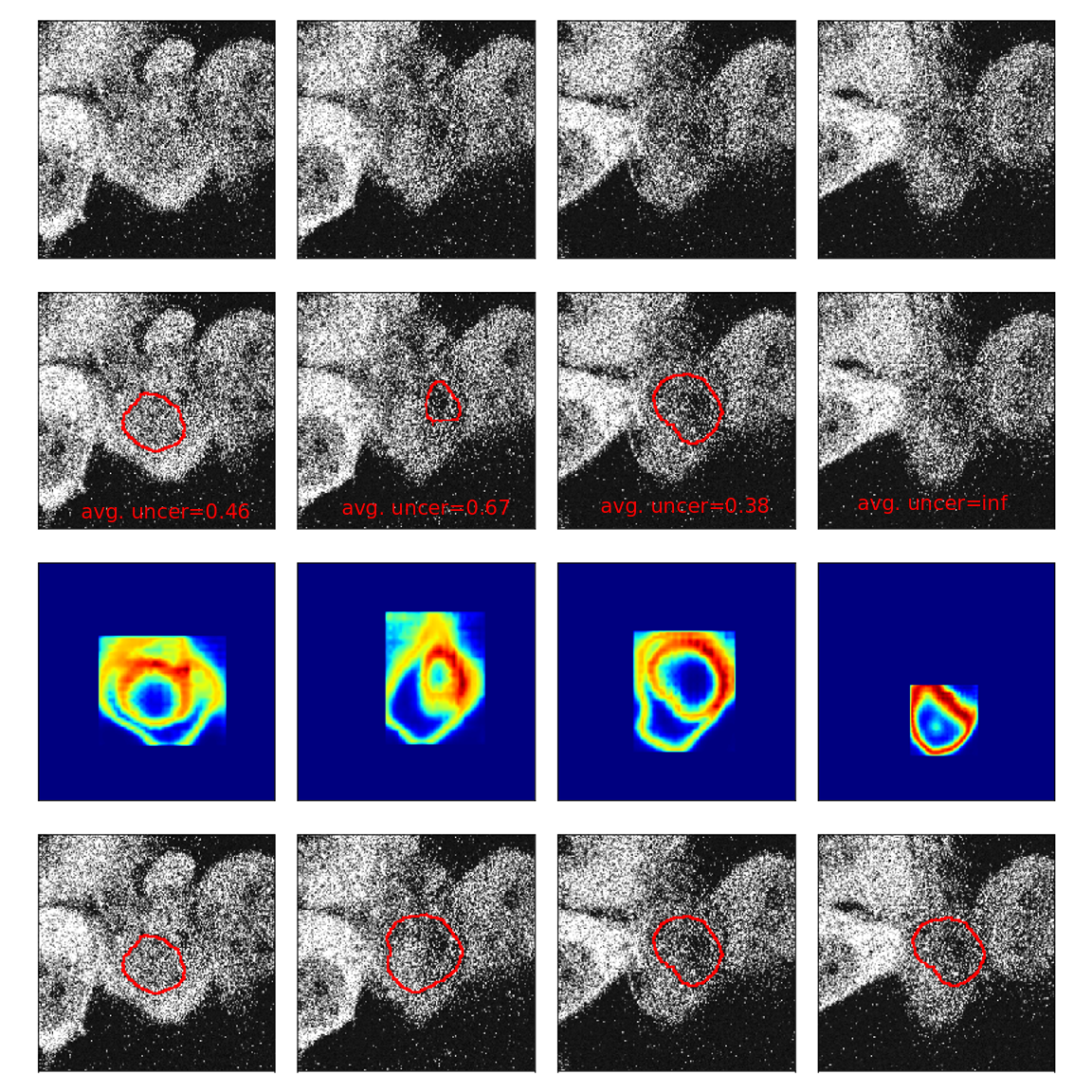}
   \caption{Example of a nucleus becoming completely indistinguishable from the surrounding cytosol in all time points. Columns correspond to time points $t-1$, $t$, $t+1$ and $t+2$, from left to right respectively. In the first row we show the raw images. In the second row we overlay MaskRCNN outputs with the input image and report the average uncertainty. Pixel-wise uncertainty maps are presented in the third row while updated masks are overlaid to the original images in the last row.}
   \label{fig:overview2}
   \vspace{-3mm}
 \end{figure}

The scenario in Figure~\ref{fig:overview2} shows the full potential of our method in the case of extreme signal loss. Uncertainty estimation allows to rank segmentation masks and infer more reasonable masks for the less certain frames. In the last frame the network misses the nucleus completely in a very confident way. Since we set the uncertainty in such cases to infinity to ensure that it gets updated (biologically there is certainly a nucleus in each cell), we are able to recover the missing nucleus. For a biologist taking measurements from the results of our method is more reliable compared to the non-updated results.
 
\subsection{No Signal Loss - Segmentation Failure}
\begin{figure}[h!]
   \centering
     \includegraphics[width=0.9\textwidth]{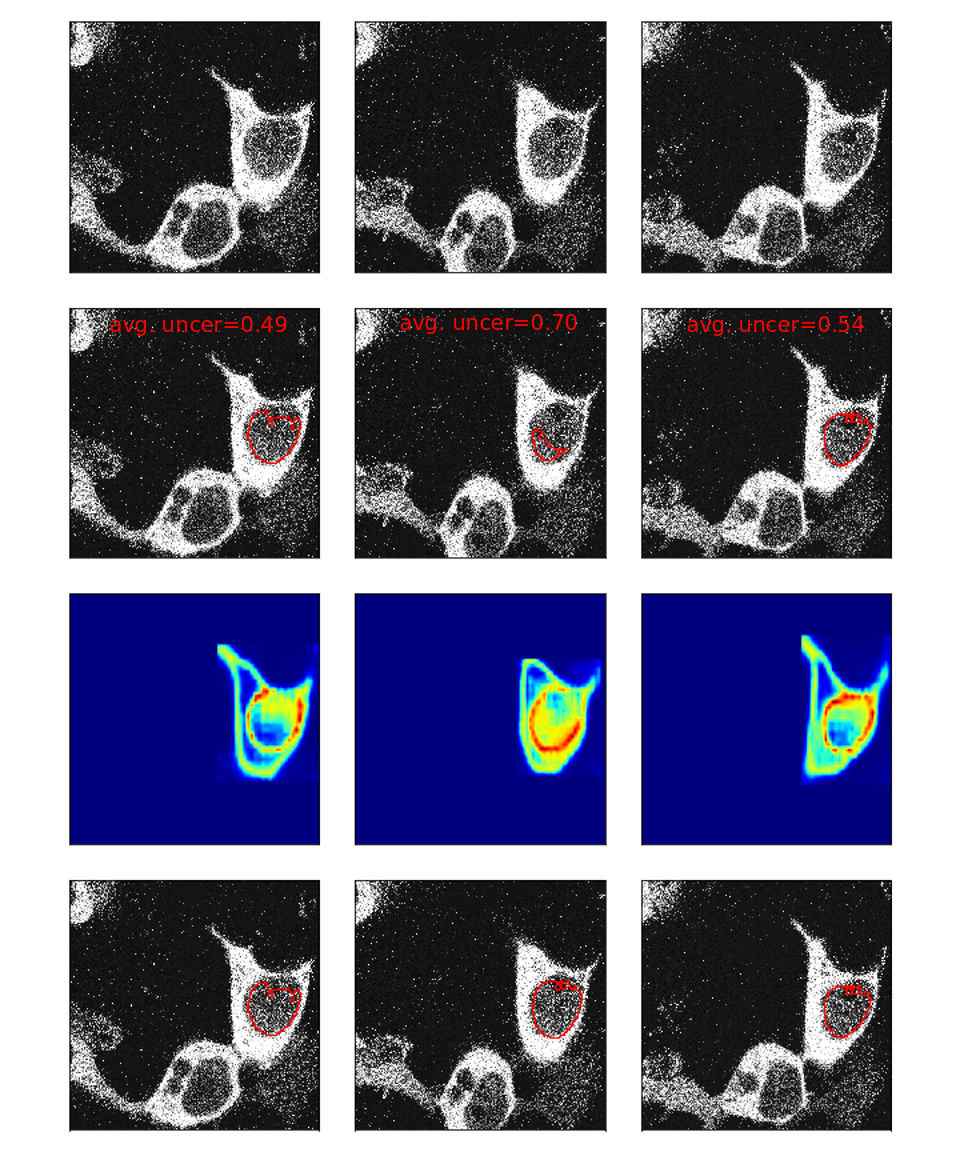}
   \caption{Example of recovery from an under-segmentation failure at time stamp $t$. Columns correspond to time points $t-1$, $t$ and $t+1$, from left to right respectively. In the first row we show the raw images. In the second row we overlay MaskRCNN outputs with the input image and report the average uncertainty. Pixel-wise uncertainty maps are presented in the third row while updated masks are overlaid to the original images in the last row.}
\label{fig:overview3}
 \end{figure}

If the training data is not rich with all variations of the real task at hand, which is often the case in biomedical settings, networks can also fail at generalization. Our method is not only limited to refining failures due to signal loss, on the contrary, it is generic to all failures caused by the uncertainties inherited in the network or the data. Our predicted uncertainty estimations provide a good indicator for these scenarios and allows to recover them as we show in Figure~\ref{fig:overview3}.   
\vspace{-0cm}
\section{Conclusion}
We addressed automated segmentation of image sequences, which cannot be analyzed frame-by-frame due to temporary uncertainties causing errors in predictions. First, we estimate uncertainty from Mask R-CNN to identify unreliable predictions. Second, we improve the less reliable predictions by propagating the more certain ones from neighbouring frames. We evaluated our method on HEK293T cells expressing a protein that oscillates in and out of the nucleus over time making the nucleus invisible temporarily. Our method improves nuclei segmentation over several baselines while keeping the cytosol and background segmentation untouched. We believe that our method will facilitate further timeseries analysis for quantitative biology to understand the effect that the dynamic localization of the protein has on the cell without additional markers.

\subsubsection*{Acknowledgments.} This project was funded by the German Research Foundation (DFG) and the German Ministry of Education and Science (BMBF). Gefördert durch die Deutsche Forschungsgemeinschaft (DFG) im Rahmen der Exzellenzstrategie des Bundes und der Länder – EXC-2189 – Projektnummer 390939984 und durch das Bundesministerium für Bildung und Forschung (BMBF) Projektnummer 01IS18042B und 031L0079.

\bibliographystyle{splncs04}
\bibliography{references}

\end{document}